\documentclass{article}
\usepackage{spconf,amsmath,graphicx}
\usepackage{subcaption}
\usepackage{csquotes}

\newcommand{\Z}{\mathbf{Z}}
\newcommand{\z}{\mathbf{z}}
\newcommand{\W}{\mathbf{W}}

\newcommand{\x}{\mathbf{x}}
\newcommand{\y}{\mathbf{y}}

\newcommand{\D}{\mathcal{D}}
\newcommand{\F}{\mathcal{F}}

\newcommand{\Obs}{\mathcal{O}}

\title{Visualizations Relevant to the User by Multi-View Latent Variable Factorization}
\name{Seppo Virtanen$^{\star}$, Homayun Afrabandpey$^{\dagger}$ and Samuel Kaski$^{\dagger}$\thanks{We thank the Academy of Finland for funding (Finnish Centre of Excellence in Computational Inference Research COIN)}}
\address{$^{\star}$Department of Statistics, University of Warwick\\
	$^{\dagger}$ Helsinki Institute for Information Technology HIIT, Department of Computer Science, Aalto University}
\begin{document}
	%
	\maketitle
	\begin{abstract}
		A main goal of data visualization is to find, from among all the available alternatives, mappings to the 2D/3D display which are relevant to the user. Assuming user interaction data, or other auxiliary data about the items or their relationships, the goal is to identify which aspects in the primary data support the user's input and, equally importantly, which aspects of the user's potentially noisy input have support in the primary data. For solving the problem, we introduce a multi-view embedding in which a latent factorization identifies which aspects in the two data views (primary data and user data) are related and which are specific to only one of them. The factorization is a generative model in which the display is parameterized as a part of the factorization and the other factors explain away the aspects not expressible in a two-dimensional display. Functioning of the model is demonstrated on several data sets.
	\end{abstract}
	\begin{keywords}
		Data visualization, latent factor models, manifold embedding,
		multi-view learning
	\end{keywords}
	\section{Introduction}
	\label{sec:intro}
	
	In the machine learning community there has been a strong trend in
	developing methods for non-linear dimensionality reduction and
	manifold embedding, that is, finding lower-dimensional manifolds
	within high-dimensional data spaces. Examples of these methods include Laplacian eigenmap \cite{belkin2003laplacian}, Isomap
	\cite{tenenbaum2000global}, locally linear embedding
	\cite{roweis2000nonlinear} and stochastic neighbor embedding
	\cite{hinton2002stochastic}. For a comparison of several methods see \cite{TiCC-TR}. Different methods aim at preserving different geometric properties, but common to all is that they produce a lower-dimensional output where similar data points are located closer together than dissimilar data points.
	
	The low-dimesional projections can be used to construct scatterplots of the data, to fulfill the constantly increasing need for data visualizations across a wide range of applications. If the dimensionality of the data manifold is two, the methods work well for visualization. If it is higher, a simple solution is to choose the output dimensionality to be two, essentially compressing the data manifold. It has turned out that most of the methods are not able to do that well, but by formulating the cost functions in terms of misses and false positives, a desired tradeoff between the two can be optimized for the visualization \cite{venna2010information}.
	
	Assuming all dimensions of the data manifold are not equally relevant to the user, a better solution is to focus on visualizing the relevant ones. In this paper we assume data are available about the user preferences, from which the visualization can be learned, but the data about preferences may be indirect. A simple example is explicit feedback about groups of data being similar, which can be expressed as cluster memberships or data classes. 
	These types of auxiliary data can contain noise that is structured in the sense of containing classes that are not visible in the primary data. 
	Given the primary data, and auxiliary data about user preferences, the task of finding the relevant aspects of the data is then essentially a two-view learning problem: identify what is statistically shared in the two data views. We additionally will want the shared aspects to be used for the visualization, requiring that the rest of the signals, ``structured noise,'' is explained away. This is what our model is capable of handling.
	
	We build on the popular stochastic neighbor embedding (SNE) method 
	to infer the visualizations
	. SNE has earlier been formulated for multiple views, as multi-view stochastic neighbor embedding (mSNE) \cite{xie2011m}. That work integrated the features into a unified representation but did not yet consider visualization. The model assumed a single set of low-dimensional latent variables which explains all views, and hence cannot directly separate the source-specific ``structured noise'' from signal. A related model called multiple relational embedding \cite{memisevic2004multiple} introduced view-specific mappings that can switch latent variables off from the views and hence could implement view-specific ``explaining away.'' They did not aim at separating shared variation from view-specific, however, and did not consider visualization yet either.
	
	%
	
	In summary, the main contribution of this paper is that we formulate a generative probabilistic model to solve the problem of learning a visualization relevant to the user, operating on two-view data. The first view consists of the primary data, and the second view of auxiliary data collected from the user. The main difference between our model and the existing models is that our model has a set of latent variables, some of which are coordinates of the visualization, and the rest explain away the parts of data not relevant to the user.
	
	\section{Model}
	\label{sec:model}
	
	%
	
	We propose a generative model for two relational count data sets, denoted as $\D$ and $\F$, that represent similarities between pairs of $N$ items. For example, the items can correspond to scientific articles
	or other documents. We denote the count between items $i$ and $j$ as $d_{i,j}$ for $\D$ and $f_{i,j}$ for $\F$, respectively. We assume the counts are symmetric, that is, $d_{i,j}=d_{j,i}$ for all $i,j\in \{1,\dots,N\}$. High count indicates strong similarity between two items, whereas low count indicates the two items are less similar. However, we do not assume that counts should be similar between the data sets.
	
	We model the data $\D$ with a distribution $p$ and the user data $\F$ with a distribution $q$, both defined over pairs of data items. 
	Assuming that user data are available for only a subset $\Obs$ of data pairs, and the data $\D$ for all pairs, the joint generative distribution of the parameters and data is
	\begin{equation}
		p(\D,\F,\Theta) \propto   \prod_{i=1,j>i}^N p_{i,j}^{d_{i,j}}  \prod_{i',j' \in \Obs} q_{i',j'}^{f_{i',j'}},
	\end{equation}
	where we have collected all parameters to $\Theta$. In the following, for brevity, we limit our notation to the data set $\D$ and variables related to it, noting that similar constructions apply for the $\F$, due to symmetry. We then normalize the observed counts to a distribution over the items denoted by $\tilde d$ and $\tilde f$, and use the mean-normalized log-likelihood.
	
	
	\subsection{Data visualization and two-view learning}
	
\begin{figure}
		\centering
		\includegraphics[width=4cm]{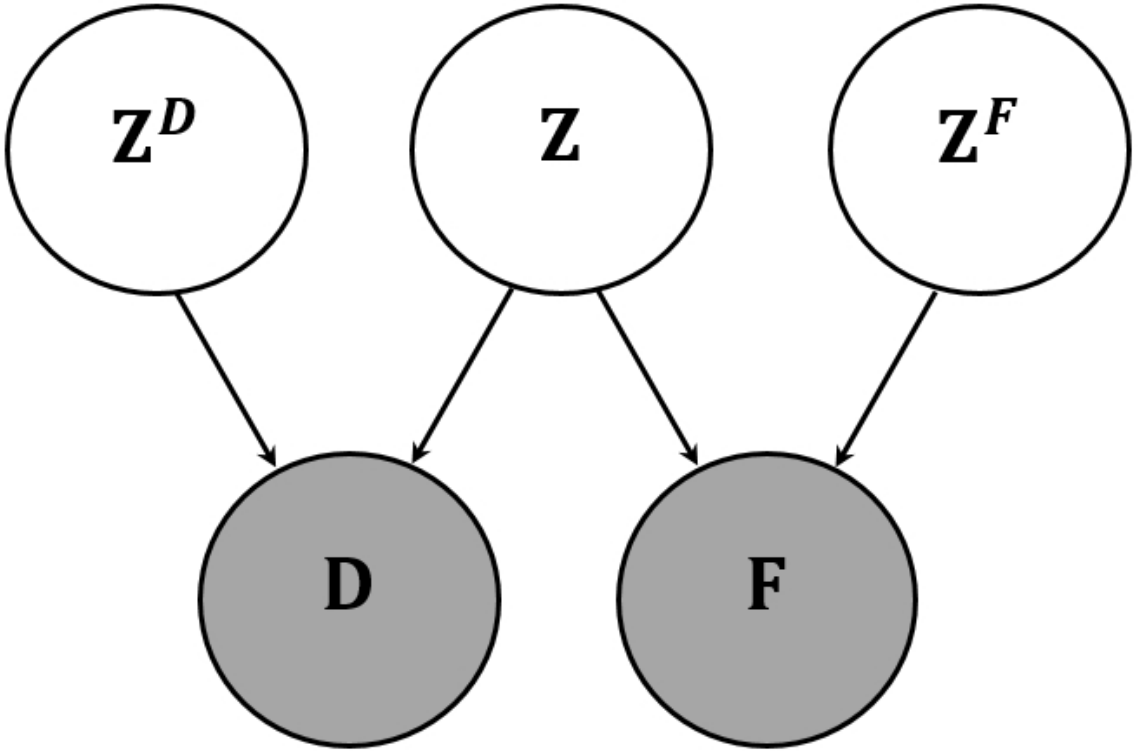}
		\caption{Graphical model for the two-view latent variable model. Gray and white nodes depict observed and hidden variables, respectively. The $\Z^{\boldsymbol{D}}$, $\Z$, and $\Z^{\boldsymbol{F}}$ are matrices containing all primary-data-specific latent variables ($\z^{\D}_i$), shared latent variables ($\z_i$), and user-data-specific latent variables ($\z^{\F}_i$), respectively. In more detail, the entry $d_{ij}$ of $\mathbf{D}$ depends on the rows $i$ and $j$ (shared vectors $\z_i$ and $\z_j$) of $\textbf{Z}$, and the rows $i$ and $j$ of $\Z^{\boldsymbol{D}}$; the dependencies for $f_{i,j}$ are analogous.}
		\label{fig1}
	\end{figure}

	For learning the visualization, we introduce three vector-valued (factorised) latent variables for each item. The variables $\z$ are shared by the two views, and the $\z^{(\D)}$ and $\z^{(\F)}$ are specific to their respective view. With these latent variables, we construct the latent (mean data) distributions as
	\begin{align}
		p_{i,j} & \propto \exp ( - || \z_i - \z_j  ||^2 - || \z^{(\D)}_i - \z^{(\D)}_j ||^2 ),
		\label{eq:fact} \\
		q_{i,j} & \propto \exp ( - || \z_i - \z_j  ||^2 - || \z^{(\F)}_i - \z^{(\F)}_j ||^2 ), \notag
	\end{align}
	where $|| \cdot ||$ denotes the Euclidean distance. The idea is that the shared latent variables capture dependencies (common joint variation) between the two data sets, 
	whereas the data set-specific latent variables capture non-shared variation for each data set. 
	Figure \ref{fig1} shows a graphical representation of our proposed model where, for simplicity, we have collected the latent variable vectors into matrices to more clearly show the main dependencies.
	
	In this parameterization, distances between the latent variables for any two items are proportional to pair-wise dissimilarity between them. Assuming a large $p_{i,j}$, the model prefers keeping the distance between latent points $i$ and $j$ small. In contrast, a small $p_{i,j}$ does not affect the cost function as strongly. This insight underlies stochastic neighbor embedding and related further developments \cite{hinton2002stochastic,venna2010information,peltonen2011generative}.
	
	In this work, we assume the shared latent variables are 2D and use them as visualisation coordinates. Assuming the view-specific latent variables have sufficient modelling flexibility to explain view-specific variation, the shared latent variables will capture the most significant commonalities between the data sets.
	
	\subsection{Group-sparse formulation}
	
	The model can be written more compactly by concatenating the latent variables together,
	\begin{equation*}
		\y_i=[\z_i,\z^{(\D)}_i,\z^{(\F)}_i] \; ,
	\end{equation*}
	and introducing data set-specific binary indicator variables $b^{(\D)}_k$ and $b^{(\F)}_k$, for $k=1,\dots,K$, that represent the latent variables either as active (one) or non-active (zero) in the corresponding data set. We note that the explicit factorization of the latent space in Equation \ref{eq:fact} is useful for understanding the structure of the multi-view model and the role the different latent variables play. However, learning with a model that has fixed factorized latent variables induces severe identifiability problems for any local learning algorithm. An approach to alleviate this problem is to assume a common latent space formulation and relax the binary indicator variables to continuous variables $\W^{(\D)}$ and $\W^{(\F)}$, respectively. Assuming sparsity for the indicator variables, during learning some of them will approach zero, shutting off the corresponding view.
	
	Thus, we re-parameterize
	\begin{equation}
		p_{i,j} \propto \exp ( - (\y_i - \y_j)^T \W^{(\D)}\W^{(\D)^T}(\y_i-\y_j) ),
	\end{equation}
	where $\y_i$ for $i=1,\dots,N$ are the (concatenated) $K$-dimensional latent variables and $\W^{(\D)}$ is a $K\times K$ diagonal matrix (we make a similar modification for $q$ with $\W^{(\F)}$).
	We set the first two elements of $\W^{(\D)}$ (and $\W^{(\F)}$) to unity capturing the shared latent variables, whereas the remaining variables on the diagonal are unconstrained. In the experiments, we show empirically on multiple data sets that the construction correctly captures the dominant shared variation between the two views via the shared latent variables. 
	
	For estimating the unobserved variables (locations on the display), we used unconstrained gradient-based optimization to find the maximum likelihood estimate.
	
	%
	%
	
	\subsection{Data setup and visualization}
	
	So far, we have not specified what the $\D$ and $\F$ correspond to. In this work, the $\D$ denotes the data the user wants to visualize. They may come directly as observations of similarity between item pairs, forming count data $d_{i,j}$, or they may come as feature vectors $\x_i$, from which the similarities are computed as $\tilde d_{i,j} = \exp(-|| \x_i - \x_j ||^2/\sigma^2)$. The $\F$ denotes data provided by the user or measured from the user (more details below); they may come directly as count data $f_{i,j}$, or computed from feature vectors $\mathbf{f}_i$ as for the data $\x_i$.
	
	The first option for the types of data the user may provide, is data about pairwise similarities of the data items. The user data can be collected in an interactive data-analysis session, whereby $f_{i,j}$ would be the number of times the items $i$ and $j$ were considered similar, or derived from categorizations or classifications: $f_{i,j}$ then is the number of classes in which both $i$ and $j$ belong. A particularly handy interactive visualization scenario is where the user indicates a set of data items being similar, which reduces to the (multiple) classification setting. The second option for user data types is that a feature vector is measured from the user during interaction, or the user provides an auxiliary feature vector $\mathbf{f}_i$ for some of the $i$. They can then be converted to $\tilde f_{i,j} = \exp(-|| \mathbf{f}_i - \mathbf{f}_j ||^2/\sigma_f^2)$.
	
	In both options, the user data are regarded as indirect evidence of what is relevant to the user in the primary data. The key assumption is that aspects of the primary data that have a statistical relationship with the user data are more relevant. The rest of the user input is not relevant to the primary data, and the rest of the primary data is not supported by the user input, and hence is likely to be less relevant to the user.

	\section{Experimental evaluation}
	
	We compare our approach to the closest available alternatives; none of them have been designed for precisely the proposed task, but they can still be applied for the task. We compare to SNE that uses only one data set (not the user data), mSNE that uses both views and assumes a single set of common latent variables, and neighbourhood component analysis (NCA) that is a supervised method which assumes classes instead of another full-blown data view. We leave out comparison to MRE because code is not available, and the method would need some further development to be applicable for information visualisation; it is not obvious which latent variables to visualise.
	
	We used three different data sets for comparison: scientific articles from \cite{peltonen2013information}, Reuters Corpus Volume 1 (RCV1), and Heart Disease data set \cite{detrano1989international} from the UCI repository \cite{Lichman:2013}. We show numerical data for all but only have space to show the visualizations on one (RCV1). We use available class information as ground truth for user interest, and simulate additional structured noise to the user data (alternative classes and unstructured noise).
	For NCA we gave the advantage of using ground truth.
	
	Performance is evaluated numerically by measuring the separability of the ground truth classes on the 2D visualization. The logic is that since we know the ground truth classes to be relevant in the sense that they inhabit the shared data space, and random errors are more likely to mix up the classes than to separate them, a better visualization separates the class distributions better. As a measure of separation we use the (leave-one-out) performance of a k-nearest neighbor classifier on the visualization.

	\subsection{Reuters Corpus Volume 1}
	
	We used a subset of RCV1-v2 corpus, first used by \cite{cai2012manifold}. The subset is a document-term matrix containing $N=9,625$ documents 
	which are divided into four categories, \enquote{C15}, \enquote{ECAT}, \enquote{GCAT}, and \enquote{MCAT}
	. For each document, feature vectors are generated by the standard TF-IDF weighting scheme.
	For details about the RCV1 corpus see \cite{lewis2004rcv1}.
	
\begin{figure}[ht!]
		\centering
		\begin{subfigure}{0.31\linewidth}
			\includegraphics[width=3cm]{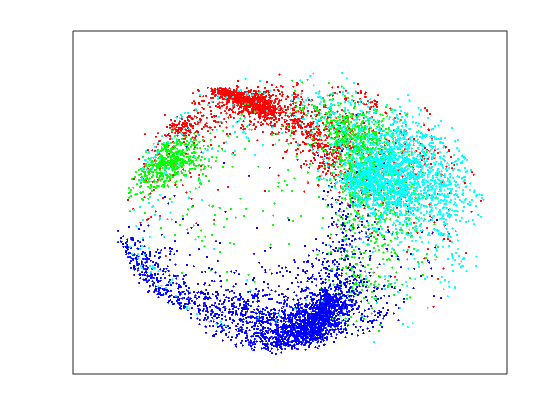}
			\caption{SNE}
			\label{fig4:sub1}
		\end{subfigure}
		\begin{subfigure}{0.31\linewidth}
			\includegraphics[width=3cm]{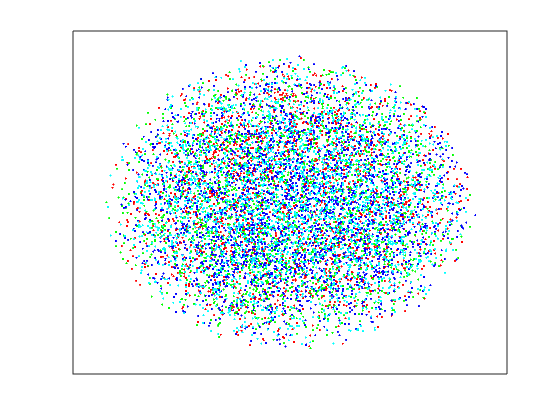}
			\caption{mSNE}
			\label{fig4:sub2}
		\end{subfigure}
		\begin{subfigure}{0.31\linewidth}
			\includegraphics[width=3cm]{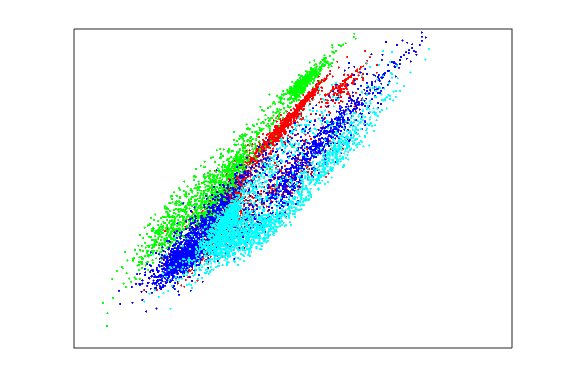}
			\caption{NCA}
			\label{fig4:sub3}
		\end{subfigure}
		\begin{subfigure}{0.31\linewidth}
			\includegraphics[width=3cm]{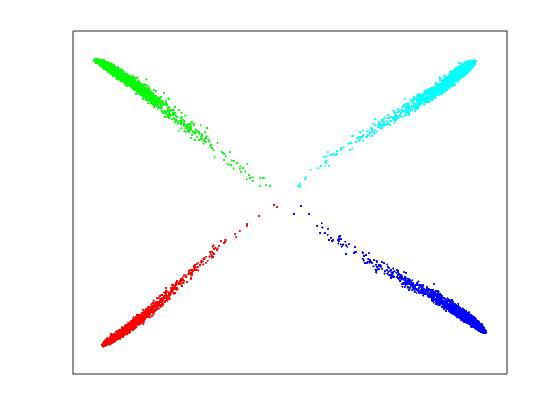}
			\caption{our model}
			\label{fig4:sub4.a}
		\end{subfigure}
		\begin{subfigure}{0.33\linewidth}
			\includegraphics[width=3cm]{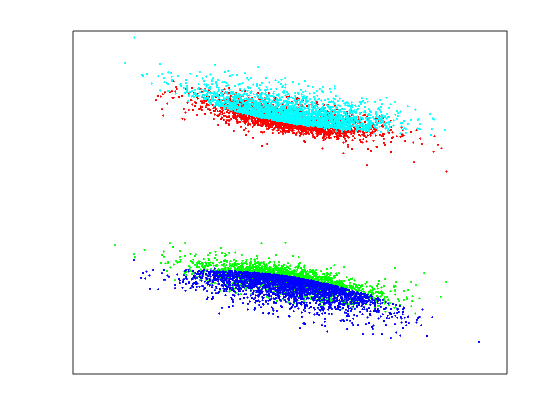}
			\caption{irrelevant user data}
			\label{fig4:sub4.b}
		\end{subfigure}
		\begin{subfigure}{0.33\linewidth}
			\includegraphics[width=3cm]{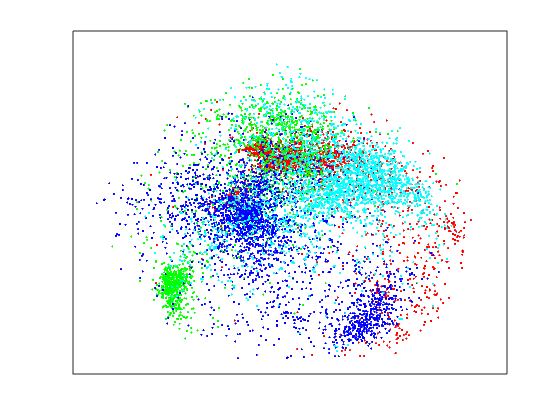}
			\caption{irrelevant data}
			\label{fig4:sub4.c}
		\end{subfigure}
		\caption{Comparison of methods in visualizing RCV1. 
			(a): SNE; (b): mSNE; (c): NCA; (d) our model. Additionally, (e) shows user-data-specific visualization for our model, and (f) the irrelevant aspects of the primary data, extracted by our model.}
		\label{fig4}
	\end{figure}

	Figure~\ref{fig4} shows our method finds the relevant structure
	clearly by inferring well-separated class-specific clusters. In this figure, categories are shown by colors of the dots: red, green, blue, and cyan represent \enquote{C15}, \enquote{ECAT}, \enquote{GCAT}, and \enquote{MCAT}, respectively. Some of the relevant structure is visible in the user-data-specific latent variables (Fig.~\ref{fig4:sub4.b}), and also in the irrelevant data (Fig.~\ref{fig4:sub4.c}), indicating that the dimensionality of the relevant structure is higher than two-dimensional and hence a higher-dimensional display would be needed to display all of the ground truth. For the alternative methods, based on the figure, we see that mSNE fails for the task, SNE infers two (noisy) clusters with incorrect classes and NCA captures a single cluster. For the other two data sets the behaviour of our proposed method was analogous and it separated the clusters well using the shared latent variables. Due to the lack of space we only show quantitative indicators on those two data sets, in Table \ref{tab1}, instead of visualizations.
	
	The classes have become separated quantitatively as well, as shown with 5-nearest neighbor classification results in the second column of Table \ref{tab1}. We can see that the proposed multi-view latent variable model outperforms other techniques with a clear margin. It may seem striking that even the supervised NCA is clearly worse; the reason is that the data contain also irrelevant classes, and NCA is not better in distinguishing the relevant and irrelevant ones. The table also contains the results for the two other data set where again our method is the best.
	
	
	\begin{table}[h!]
		\centering
		\includegraphics[width=0.9\linewidth]{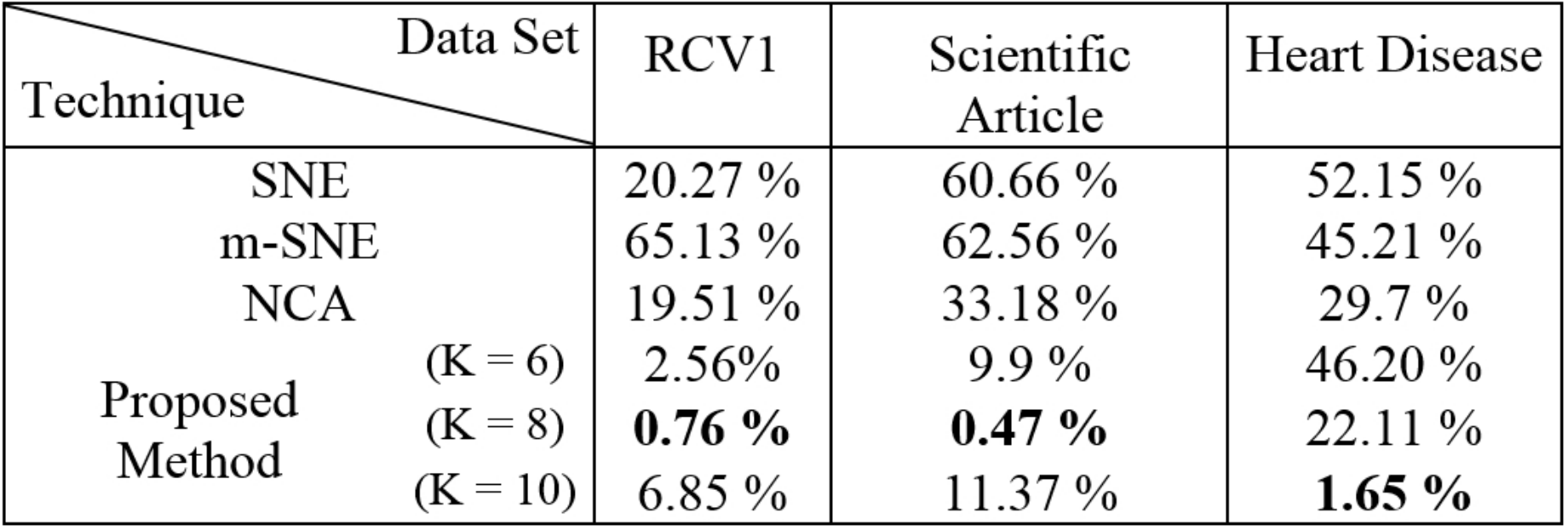}
		\caption{Quality of visualizations measured by separability of ground truth class distributions in the visualization, measured by a leave-one-out 5-NN classifier. The best result for each data set is marked in bold font. To evaluate sensitivity of our method to the number of components (of which $K-2$ are used to explain away irrelevant variation), results are shown for three different values of $K$.}
		\label{tab1}
	\end{table}
	\subsection{Multi-labelling}
	
	To demonstrate that different aspects of the same data set can be visualized for different users, according to what is relevant to them, we simulate two users who are interested in different aspects and give different feedbacks (labellings). We used the Abalone data set from the UCI repository. This is a classification data set aimed at predicting the age of abalones from physical measurements. The number of rings inside the shell is a significant factor for determining the age of an abalone. We thresholded the numbers into three categories where the first group contains 3 to 9 rings, the second group contains 10 to 16 rings and finally the last group contains 17 to 23 rings. We left out ring numbers having less than 5 samples.
	This resulted in only 9 discarded records, which is negligible in the total number of 4177.
	
\begin{figure}[ht!]
		\centering
		\begin{subfigure}{0.48\linewidth}
			\includegraphics[width=3.5cm]{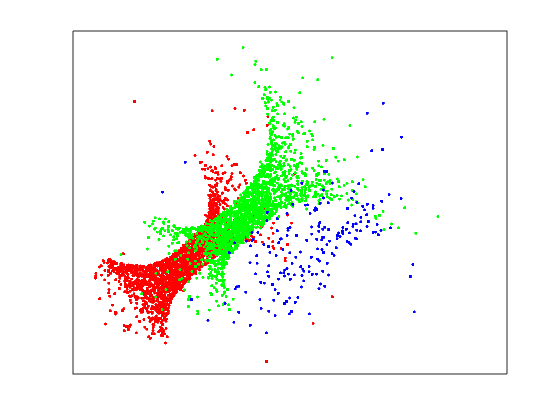}
			\caption{Age}
			\label{Fig6:sub1}
		\end{subfigure}
		\begin{subfigure}{0.48\linewidth}
			\includegraphics[width=3.5cm]{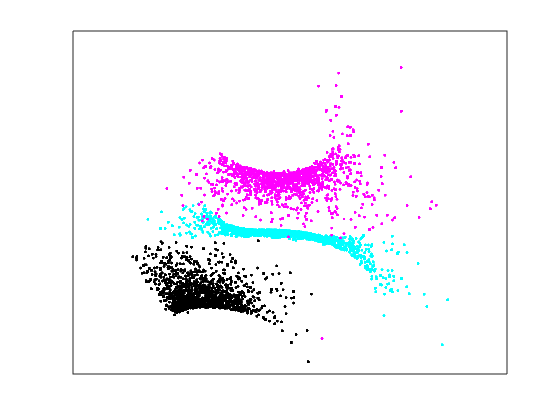}
			\caption{Sex}
			\label{Fig6:sub2}
		\end{subfigure}
		\caption{Visualization of the Abalone data for two users
			interested in different aspects of the data. (a) The colors correspond to different age groups (red: group 1; green: group 2, blue: group 3). (b) Cyan: M; magenta: F; black: I.}
		\label{fig6}
	\end{figure}

	The first user is interested in the age and assigns similarities according to ring numbers. The second user is interested in sex and gives feedback according to the three categories M, F, and I. Figure \ref{fig6} shows the visualization of the shared latent
	variables of our method for the two different labellings. In
	\ref{Fig6:sub1} the ages become separated to an extent, and in 
	\ref{Fig6:sub2} the sexes. The visualizations are different, as they should be for two users interested in different aspects, both having support in the data.
	
	\section{Conclusion}
	
	We have introduced a statistical principle to identify and visualize aspects of data relevant to the user, by exploiting statistical relations found between the primary data, and user-provided auxiliary data. Unlike manifold embedding-based dimensionality reduction methods which have not been designed for compressing dimensionalities to two for visualization on display, our proposed method is able to visualize well in two dimensions, by explaining away the irrelevant data with additional latent variables.
	
	In this paper, we successfully demonstrated and compared the proposed method on multiple static data sets, where the ground truth came from categorizations of the data. A main future goal is to use similar techniques in interactive visualization, where user interaction data will be measured all the time, and the visualization needs to react faster. 
	
	We did not consider model order selection in this paper. Our
	expectation is that standard probabilistic methods, in particular automatic relevance determination (ARD), could be used for determining the total number of latent variables
	. It may turn out to be more difficult to choose how many latent variables are relevant for the user, in case the relevant subspace cannot be adequately presented in 2D. Our current hypothesis is that group sparsity combined with ARD would be sufficient.
	
	
	
	%

	\bibliographystyle{plain}
	\bibliography{Visualizations_Relevant_to_The_User_By_Multi-View_Latent_Variable_Factorization.bbl}
	
\end{document}